%% file: main.tex
\documentclass[10pt,twocolumn,letterpaper]{article}

\usepackage{cvpr}              
\usepackage[accsupp]{axessibility} 
\input{preamble}

\definecolor{cvprblue}{rgb}{0.21,0.49,0.74}
\usepackage[pagebackref,breaklinks,colorlinks,citecolor=cvprblue]{hyperref}

\usepackage{xcolor}

\def\paperID{15372} 
\def\confName{CVPR}
\def\confYear{2024}

\title{You Only Need Less Attention at Each Stage in Vision Transformers}

\author{Shuoxi Zhang\textsuperscript{1}
, Hanpeng Liu\textsuperscript{1}\thanks{Equal contribution}
, Stephen Lin\textsuperscript{2}, Kun He\textsuperscript{1}\thanks{Corresponding author} \\
{\tt\small \{zhangshuoxi, hanpengliu, brooklet60\}@hust.edu.cn}, {\tt\small stevelin@microsoft.com}\\
\textsuperscript{1}Huazhong University of Science and Technology, \textsuperscript{2}Microsoft Research Asia\\
}

\begin{document}
\maketitle
\input{sec/0Abstract}    
\input{sec/1Intro}
\input{sec/2RW}

\input{sec/3Method}

\input{sec/4Exp}

\input{sec/5Con}
\newpage
{
    \small
    \bibliographystyle{ieeenat_fullname}
    \bibliography{main}
}


\end{document}

%% file: preamble.tex
\usepackage[dvipsnames]{xcolor}


%% file: sec/0Abstract.tex
\begin{abstract}
The advent of Vision Transformers (ViTs) marks a substantial paradigm shift in the realm of computer vision. ViTs capture the global information of images through self-attention modules, which perform dot product computations among patchified image tokens. While self-attention modules empower ViTs to capture long-range dependencies, the computational complexity grows quadratically with the number of tokens, which is a major hindrance to the practical application of ViTs. Moreover, the self-attention mechanism in deep ViTs is also susceptible to the attention saturation issue.  
Accordingly, we argue against the necessity of computing the attention scores in every layer, and 
we propose the Less-Attention  Vision Transformer (\name), which computes only a few attention operations at each stage and calculates the subsequent feature alignments in other layers via attention transformations that leverage the previously calculated attention scores. This novel approach can mitigate two primary issues plaguing traditional self-attention modules: the heavy computational burden and attention saturation. Our proposed architecture offers superior efficiency and ease of implementation, merely requiring matrix multiplications that are highly optimized in contemporary deep learning frameworks. Moreover, our architecture demonstrates exceptional performance across various vision tasks including classification, detection and segmentation.
\end{abstract}

%% file: sec/1Intro.tex
\section{Introduction}
\label{sec:intro}

In recent years, computer vision has experienced rapid growth and development, primarily owing to the advances in deep learning and the accessibility of large datasets. Among the prominent deep learning techniques, Convolutional Neural Networks (CNNs)~\cite{krizhevsky2017imagenet} have proven particularly effective, demonstrating exceptional performance across a wide range of applications, including image classification~\cite{krizhevsky2017imagenet,wang2017residual}, object detection~\cite{girshick2015region, redmon2016you}, and semantic segmentation~\cite{badrinarayanan2017segnet,unet}.

\begin{figure*}[t]
\vspace{-30pt}
\centering
\includegraphics[width=0.80\textwidth]{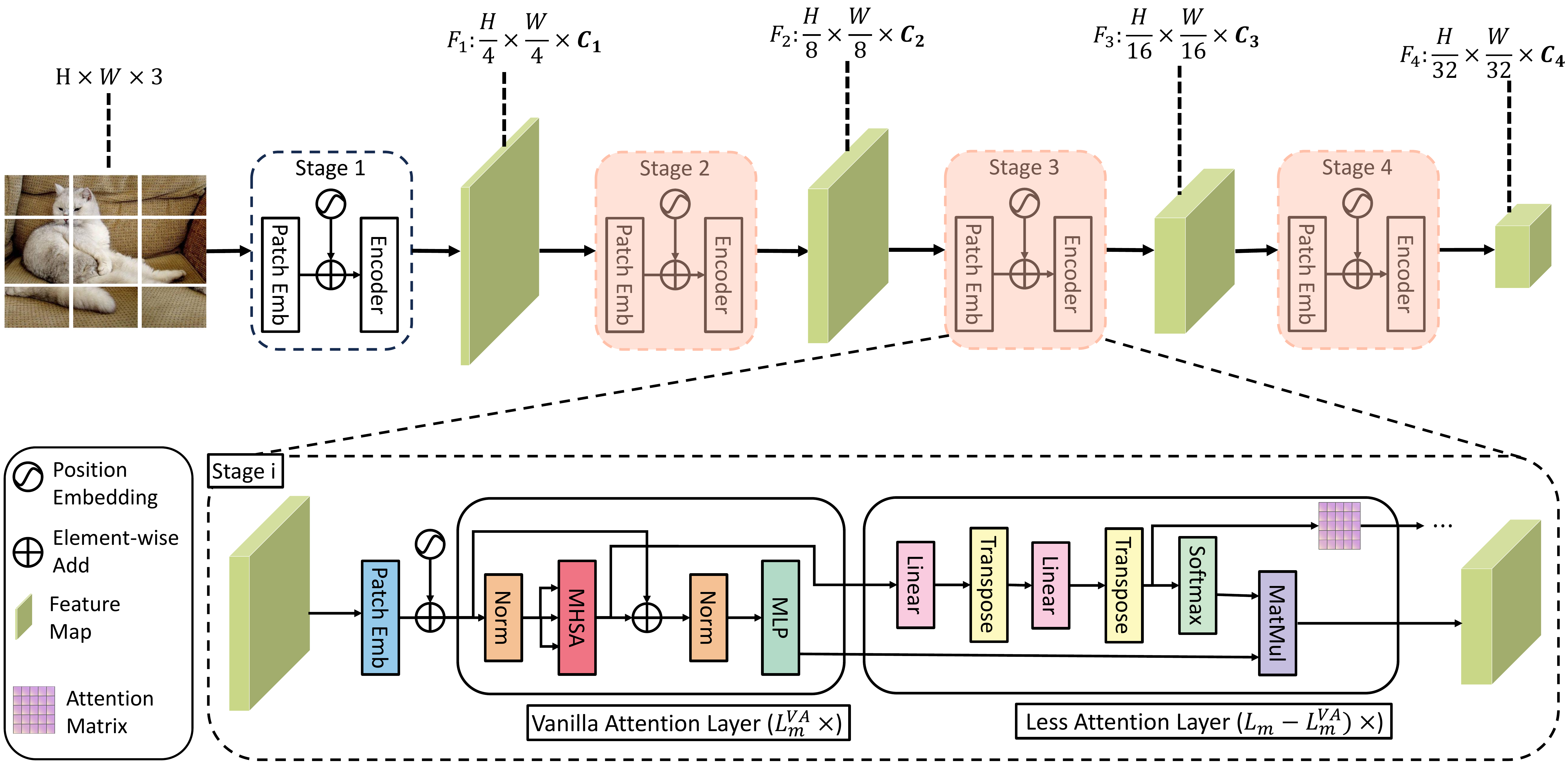} 
\caption{The architecture of our \textbf{L}ess-\textbf{A}ttention \textbf{Vi}sion \textbf{T}ransformer (\name). The bottom part: the proposed \less, which together with conventional Transformer blocks in the preceding layers constitutes the feature extraction module of this stage.}
\label{fig2}
\end{figure*}

Inspired by the great success of Transformers~\cite{vaswani2017attention} in natural language processing, Vision Transformers (ViTs)~\cite{ViT} divide each image into a set of tokens. 
These tokens are then encoded to produce an attention matrix that serves as a fundamental component of the self-attention mechanism. 
The computational complexity of the self-attention mechanism grows quadratically with the number of tokens, and the computational burden becomes heavier with higher-resolution images. 
Some researchers attempt to reduce token redundancy through dynamic selection~\cite{dynamicvit,Evit} or token pruning~\cite{wu2023ppt} to alleviate the computational burden of the attention computation. These approaches have demonstrated comparable performance to the standard ViT. However, methods involving token reduction and pruning necessitate meticulous design of the token selection module and may result in the inadvertent loss of critical tokens. In this work, we 
explore a different direction and rethink the mechanism of self-attention. In the attention saturation problem 
raised in \cite{zhou2021deepvit}, as the layers of ViTs are progressively deepened, the attention matrix tends to remain largely unaltered, mirroring the weight allocation observed in the preceding layers. Taking these considerations into account, we are prompted to pose the following question: 

\textit{
Is it really necessary 
to consistently apply the self-attention mechanism throughout each stage of the network,
from inception to conclusion?}

In this paper, we propose to modify the fundamental architecture of standard ViT by introducing the \textbf{L}ess-\textbf{A}ttention \textbf{Vi}sion \textbf{T}ransformer (\name). Our framework, as depicted in Fig.~\ref{fig2}, consists of Vanilla Attention (VA) layers and our proposed Less Attention (LA) layers to capture the long-range relationships.
In each stage, we exclusively compute the traditional self-attention and store the attention scores in a few initial Vanilla Attention (VA) layers. In subsequent layers, we efficiently generate attention scores by utilizing the previously calculated attention matrices, thereby mitigating the quadratic computational expense associated with self-attention mechanisms. Moreover, we integrate residual connections within the attention layers during downsampling across stages, allowing for the preservation of crucial semantic information learned in earlier stages while still transmitting global contextual information through alternate pathways. Finally, we carefully design a novel loss to preserve the diagonality of attention matrices during the transformation process. These key components 
enable our 
proposed ViT model to diminish both computational complexity and attention saturation, ultimately leading to notable performance improvements with reduced floating-point operations per second (FLOPs) and considerable throughput. 

To verify the effectiveness of our proposed approach, we conduct comprehensive experiments on various benchmark datasets, comparing the performance of our model with existing state-of-the-art ViT variants (also recent efficient ViTs). The experimental results demonstrate the efficacy of our approach in addressing attention saturation and achieving superior performance in visual recognition tasks.

Our main contributions are summarized as follows:
\begin{itemize}
\setlength{\itemsep}{0pt}
\setlength{\parsep}{0pt}
\setlength{\parskip}{0pt}
    \item We present a novel ViT architecture that generates attention scores by re-parameterizing the attention matrix computed by preceding layers. This approach addresses both the attention saturation and the associated computational burden.
    
    \item Moreover, we propose a novel loss function that endeavors to preserve the diagonality of attention matrices during the process of attention re-parameterization. We posit that this is essential to uphold the semantic integrity of attention, ensuring that the attention matrices accurately reflect the relative importance among input tokens.
    
    \item Our architecture consistently performs favorably against several state-of-the-art ViTs, while having similar or even reduced computational complexity and memory consumption, across various vision tasks including classification, detection and  segmentation.
\end{itemize}



%% file: sec/2RW.tex
\section{Related Work}

\subsection{Vision Transformers}
The Transformer architecture, initially introduced for machine translation~\cite{vaswani2017attention}, has since been applied to computer vision tasks through the growth of ViT~\cite{ViT}. The key innovation of ViT lies in its capability to capture long-range dependencies between distant regions of the image, achieved through the incorporation of self-attention mechanisms. 

Drawn from the triumph of ViT, a plethora of variant models have emerged, each devised to ameliorate specific constraints inherent to the original architecture. For instance, DeiT~\cite{touvron2021training} enhances data efficiency during training by incorporating the distillation token. Additionally, CvT~\cite{wu2021cvt} and CeiT~\cite{yuan2021incorporating} integrate the convolutional structure into the ViT framework to combine the strengths of CNNs (spatial invariance) and ViTs (long-range dependency modeling). These advancements underscore the ongoing evolution of transformer-based architectures in the field of computer vision.

\subsection{Efficient Vision Transformers}
Though highly effective, ViTs suffer from a huge computational burden. The research on efficient vision transformers addresses the quadratic cost of the self-attention operation by including hierarchical downsampling operations~\cite{wang2021pyramid,wang2022pvt,pan2021hvt}, token reduction~\cite{Evit,dynamicvit, wu2023ppt}, or lightweight architectural designs~\cite{pan2022less, pan2022fast}.
Hierarchical downsampling operations address the quadratic computation of self-attention by reducing the token numbers gradually across the stages~\cite{pan2021hvt, pan2022fast, wang2021pyramid, wang2022pvt} and enable ViTs to learn hierarchical structures. Another research direction introduces the token selection module to eliminate the least meaningful tokens and reduce the computational burden. For instance, \cite{dynamicvit, Evit, wu2023ppt}  reorganize image tokens by preserving informative image tokens and dropping ones with little attention to expedite subsequent MHSA and FFN computations.




\subsection{Attention Mechanisms}
The key component of 
ViTs is the attention mechanism, which computes pairwise interactions between all patches, resulting in quadratic complexity with respect to the input size. This problem leads to heavy inference computation, which hinders the practical application of ViTs in the real world.  
Several studies argue that the computational burden can be alleviated by utilizing sparse attention mechanisms, which selectively attend to a subset of patches based on their relevance or proximity. One notable approach is the Adaptive Sparse Token Pruning framework~\cite{wei2023sparsifiner}, which induces a sparse attention matrix, effectively addressing computational efficiency concerns. Furthermore, employing techniques like structured sparsity patterns~\cite{chen2021chasing,han2021demystifying,dynamicvit} can further reduce computational complexity, thereby enhancing the overall efficiency of ViTs.
Another urgent issue to be addressed 
is the problem of attention saturation, where the attention matrix displays limited variation as the layer depth increases. This issue has been acknowledged in studies such as DeepViT~\cite{zhou2021deepvit} and CaiT~\cite{cait}, which report that attention saturation hinders the ability of deep ViTs to capture additional semantic information and may even reduce training stability. Therefore, it is essential to carefully design the self-attention mechanism in 
ViTs to avoid sub-optimal solutions. 

%% file: sec/3Method.tex
\section{Methodology}
\begin{figure}[t]
\centering
\vspace{-60pt}
\includegraphics[width=0.85\columnwidth]{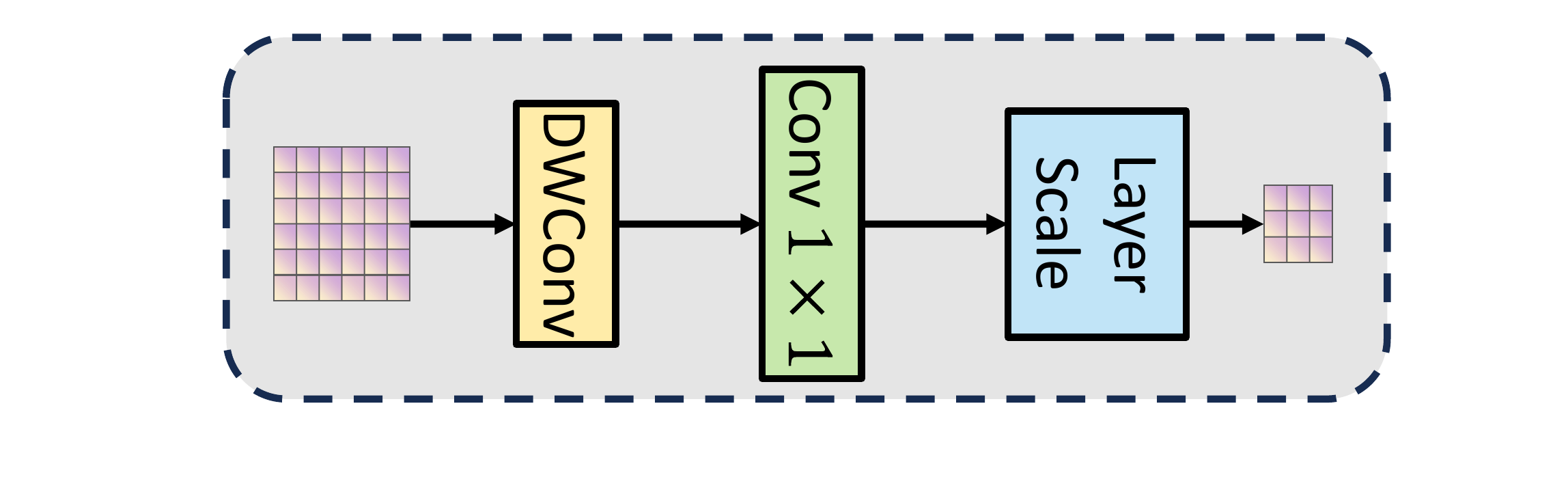} 
\caption{
The downsampling on attention across stages.
}
\vspace{-10pt}
\label{fig1}
\end{figure}

In this section, we first review the basic design of the hierarchical vision transformer. Then, we discuss two major weaknesses of its attention mechanism, and propose that self-attention can be grasped with \textit{less attention at each stage}. We dynamically re-parameterize the attention scores by utilizing the stored attention matrix from the previous layer, effectively mitigating the issue of attention saturation. Additionally, we integrate residual connections to facilitate the transfer of global relationships from earlier stages. Last but not least, we introduce a novel loss, the Diagonality Preserving loss, to preserve basic properties in the transformed attention (\ie, representing the relationships among tokens).

\subsection{Vision Transformer}
\label{sec: prelim}
Let $\mathbf{x} \in \mathbb{R}^{H \times W \times C}$ 
represent an input image, where $H \times W$ denotes the spatial resolution and $C$ the number of channels. We first tokenize the image by partitioning it into $N = \nicefrac{HW}{p^{2}} $ patches, where each patch $P_i \in \mathbb{R}^{p \times p \times C}\left(i \in \{1, \ldots, N\} \right)$ has a size of $p \times p$ pixels and $C$ channels. The patch size $p$ is a hyper-parameter that determines the granularity of the token.
The patch embedding can be extracted by using a convolution operator with the stride and kernel size equal to the patch size. Each patch is then projected to the embedding space $\boldsymbol{Z} \in \mathbb{R}^{N\times{D}}$ through non-overlapping convolution, where $D$ represents the dimension of each patch.

{\noindent \bf Multi-Head Self-Attention.} \quad
 We first provide a brief overview on the vanilla self-attention mechanism that processes the embedded patches and functions within the framework of Multi-Head Self-Attention blocks (MHSAs).
In the $l$-th MHSA block, the input $\boldsymbol{Z}_{l-1}, l \in \{1,\cdots, L\}$, is projected into three learnable embeddings  $\{\mathbf{Q,K,V}\} \in \mathbb{R}^{N \times D}$.  The multi-head attention aims to capture the attention from different views; for simplicity, we choose $H$ heads, where each head is a matrix with the dimension $N \times \frac{D}{H}$.  The $h$-th head attention matrix $\mathbf{A}_h$ can be calculated by:
\begin{align}
    \mathbf{A}_h =
    \mathrm{Softmax} \left(\frac{\mathbf{Q}_h \mathbf{K}_h^\mathsf{T}}{\sqrt{d}} \right) \in \mathbb{R}^{N \times N}.
    \label{eq:attn}
\end{align}
$\mathbf{A}_h, \mathbf{Q}_h$, and $\mathbf{K}_h$ are the attention matrix, query, and key of the $h$-th head, respectively. We also split the value $\mathbf{V}$ into $H$ heads. To avoid vanishing gradients caused by the sharpness of the probability distribution, we divide the inner product of $\mathbf{Q}_h$ and $\mathbf{K}_h$ by $\sqrt{d}$ 
($d = D/H$). The attention matrix is concatenated as:
\begin{equation}
\begin{split}
    \mathbf{A} &= \textrm{Concat}(\mathbf{A}_1, \cdots, \mathbf{A}_h, \cdots,\mathbf{A}_H); \\
    \mathbf{V} &= \textrm{Concat}(\mathbf{V}_1, \cdots, \mathbf{V}_h, \cdots,\mathbf{V}_H).
\end{split}
\label{eq:concat}
\end{equation}
The calculated attention among spatially split tokens may guide the model to focus on the most valuable tokens within the visual data. Subsequently, weighted linear aggregation is applied to the corresponding value $\mathbf{V}$:
\begin{align}
    \boldsymbol{Z}^{\textrm{MHSA}} = \mathbf{AV} \in \mathbb{R}^{N \times D}.
    \label{eq:val-feats}
\end{align}

{\noindent \bf Downsampling Operation.} \quad
Several studies have incorporated hierarchical structures into ViTs, drawing inspiration from the success of hierarchical architectures in CNNs. These works partition the Transformer blocks into $M$ stages and apply downsampling operations before each Transformer stage, thereby reducing the sequence length. In our study, we employ a downsampling operation using a convolutional layer with the kernel size and stride set to $2$. This approach permits the flexible adjustment of the feature map's scale at each stage, thereby establishing a Transformer hierarchical structure that mirrors the organization of the human visual system.

\subsection{The Less-Attention Framework}


The overall framework of our network architecture is illustrated in Fig.~\ref{fig2}. In each stage, we extract the feature representation in two phases. At the initial several Vanilla Attention (VA) layers, we conduct the standard MHSA operation to capture the overall long-range dependencies. Subsequently, we simulate the attention matrices to mitigate quadratic computation and address attention saturation at the following Less-Attention (LA) layers by applying a linear transformation to the stored attention scores. Herein, we denote the attention score before the $\textrm{Softmax}$ function of the initial $l$-th VA layer in the $m$-th stage as $\mathbf{A}^{\text{VA},l}_m$, which is computed by the following standard procedure:
\begin{equation}
    \mathbf{A}^{\text{VA},l}_m = \frac{\mathbf{Q}^l_m(\mathbf{K}^l_m)^\mathsf{T}}{\sqrt{d}}, ~~ l \leq L^{\text{VA}}_m.
\label{eq:init}
\end{equation}
Here, $\mathbf{Q}_m^l$ and $\mathbf{K}_m^l$ represent the queries and keys from the $l$-th layer of the $m$-th stage, following the downsampling from the preceding stage. And $L^{\text{VA}}_m$ is used to denote the number of VA layers. After the initial vanilla attention phase, we discard the traditional quadratic MHSA and apply transformations on $\mathbf{A}^\textrm{VA}_m$ to 
lessen the amount of attention computation. This process entails two linear transformations with a matrix transposition operation in between. To illustrate, let us consider the attention matrix in the $l$-th ($l > L^{\text{VA}}_m$) layer (LA layer) of the stage:
\begin{equation}
\begin{aligned}
    &\mathbf{A}^{l}_m = \Psi(\Theta(\mathbf{A}^{l-1}_m)^\mathsf{T})^\mathsf{T},  ~~ L^{\text{VA}}_m<l \leq L_m,\\
    &\mathbf{Z}^{\text{LA},l} = \textrm{Softmax}(\mathbf{A}^l_m)\mathbf{V}^l.
\end{aligned}
\end{equation} 
In this context, the transformations denoted by $\Psi$ and $\Theta$ refer to linear transformation layers with a dimension of $\mathbb{R}^{N\times{N}}$. Here, $L_m$, $L_m^{\text{VA}}$ represent the number of layers and the number of VA layers in the $m$-th stage, respectively. The insertion of the transposition operation between these two linear layers serves the purpose of maintaining the matrix similarity behavior. This step is essential due to the fact that the linear transformation in a single layer conducts the transformations row-wise, which could potentially result in the loss of diagonal characteristics.

\subsection{Residual-based Attention Downsampling}
When the computation traverses across stages in hierarchical ViTs, downsampling operations are often employed on the feature map. While this technique reduces the token numbers, it may result in the loss of essential contextual information. Consequently, we posit that the attention affinity learned from the preceding stage could prove advantageous for the current stage in capturing more intricate global relationships. Drawing inspiration from ResNet~\cite{he2016deep}, which introduces shortcut connections to mitigate feature saturation issues, we adopt a similar concept and incorporate it into the downsampling attention computation within our architecture. By introducing a shortcut connection, we can introduce the inherent bias into the current MHSA block. This allows the attention matrix from the previous stage to effectively guide the attention computation of the current stage, thereby preserving crucial contextual information.

However, directly applying the shortcut connection to the attention matrix might pose challenges in this context, primarily due to the difference in the attention dimensions between the current stage and the preceding stage. Here, we design an Attention Residual (AR) module that consists of a depth-wise convolution (DWConv) and a $\textrm{Conv}_{1\times1}$ layer to downsample the attention map from the previous stage while keeping the semantic information. We denote the last attention matrix (at the $L_{m-1}$ layer) of the previous stage (the $m-1$-th stage) as $\textbf{A}_{m-1}^{\text{last}}$, and the downsampled initial attention matrix of the current (the $m$-th) stage as $\textbf{A}_m^\text{init}$. $\textbf{A}_{m-1}^{\text{last}}$ has the dimension of $\mathbb{R}^{B\times{H}\times{N_{m-1}}\times{N_{m-1}}}$ ($N_{m-1}$ denotes the token number at the $m-1$-th stage). We view the multi-head dimension $H$ as the channel dimension in regular image space, thus with the $\textrm{DWConv}$ operator ($\textrm{stride}=2,\ \textrm{kernel size}=2$), we may capture the spatial dependencies among the tokens during attention downsampling. The output matrix after the $\textrm{DWConv}$ transformation fits the size of the attention matrix of the current stage, \ie, \,$\mathbb{R}^{B\times{H}\times{N_m}\times{N_m}} (N_m = \nicefrac{N_{m-1}}{2}\ \textrm{in our case})$. After depth-width convolution on the attention matrix, we then perform $\text{Conv}_{1\times1}$ to exchange information across different heads. Our attention downsampling is illustrated in Fig.~\ref{fig1}, and the transformation from $\textbf{A}_{m-1}^\text{last}$ to $\textbf{A}_{m}^\text{init}$ can be expressed as:
\begin{align}
     \textbf{A}^\textrm{init}_m &= \textrm{Conv}_{1\times1}\left(\textrm{Norm}(\textrm{DWConv}(\textbf{A}^\textrm{last}_{m-1}))\right), \label{eq:residual}
     \\
     \mathbf{A}^{\text{VA}}_m &\gets \mathbf{A}^{\text{VA}}_m + \textrm{LS}(\textbf{A}^\textrm{init}_m) \label{eq:plus},
\end{align}
where $\textrm{LS}$ is the Layer-Scale operator introduced in \cite{cait} to alleviate attention saturation. $\mathbf{A}^{\text{VA}}_m$ is the attention score for the first layer in the $m$-th stage, which is calculated by adding the standard MHSA with Eq. \ref{eq:init} and the residual calculated by Eq.~\ref{eq:residual}.

Two fundamental design principles guide our attention downsampling module. First, we utilize $\text{DWConv}$ to capture spatial local relationships during downsampling, thereby enabling the efficient compression of attention relationships. Second, the $\textrm{Conv}_{1\times1}$ operation is utilized to exchange the attention information across heads. This design is pivotal as it facilitates the efficient propagation of attention from the preceding stage to the subsequent stage. 
Incorporating the residual attention mechanism necessitates only minor adjustments, typically involving adding a few lines of code to the existing ViT backbone. It is worth emphasizing that such a technique can be seamlessly applied to various versions of the Transformer architecture. The only prerequisite is to store the attention scores from the previous layer and establish the skip-connections to this layer accordingly. The importance of this module will be further illuminated through comprehensive ablation studies.

\subsection{Diagonality Preserving Loss}
We have carefully designed the Transformer modules by incorporating attention transformation operators, aiming to mitigate the issues of computational cost and attention saturation. However, a pressing challenge remains---ensuring that the transformed attention preserves the inter-token relationships. It is well established that applying transformations to attention matrices can compromise their capacity to capture similarities, largely due to the linear transformation treating the attention matrix row-wise. Thus, we design a an alternative approach to guarantee that the transformed attention matrix retains the fundamental properties necessary to convey associations among tokens. A conventional attention matrix should possess the following two properties, \ie, diagonality and symmetry:
\begin{equation}
\begin{aligned}[b]
\mathbf{A}_{ij} &= \mathbf{A}_{ji},  \\
 \mathbf{A}_{ii} &> \mathbf{A}_{ij}, \forall j \neq i.
\end{aligned}
\label{eq:property}
\end{equation}
Thus, we design the diagonality preserving loss of the $l$-th layer to keep these two basic properties as:
\begin{equation}
\begin{split}
    {\mathcal{L}_{\textrm{DP},l}} &= \sum_{i=1}^N\sum_{j=1}^N\left|\mathbf{A}_{ij} -\mathbf{A}_{ji}\right| \\
    &+ \sum_{i=1}^N((N-1)\mathbf{A}_{ii}-\sum_{j\neq i}\mathbf{A}_{j}).
\end{split}
\end{equation}
Here, $\mathcal{L}_\textrm{DP}$ is the \textbf{D}iagonality \textbf{P}reserving loss aiming at preserving the properties of attention matrix of Eq.~\ref{eq:property}. We add our Diagonality Preserving Loss on all transformation layers with the vanilla cross-entropy (CE) loss~\cite{ViT}, thus the total loss in our training can be presented as:
\begin{equation}
\begin{aligned}[b]
    \mathcal{L}_\textrm{total} &= \mathcal{L}_\textrm{CE} + \sum_{m=1}^M\sum_{l=1}^{L_m}\mathcal{L}_{\textrm{DP},l}, \\
    \mathcal{L}_\textrm{CE} &= \textrm{cross-entropy}(Z_\texttt{Cls}, y),
\end{aligned}
\end{equation}
where $Z_\texttt{Cls}$ is the classification token of the  representation
in the last layer.

\subsection{Complexity Analysis}
Our architecture consists of four stages, each comprising $L_m$ layers. The downsampling layer is applied between each consecutive stage. As such, the computational complexity of traditional self-attention is $\mathcal{O}(N_m^2{D})$, whereas the associated K-Q-V transformation incurs a complexity of $\mathcal{O}(3N_mD^2)$. In contrast, our method leverages an $N_m\times N_m$ linear transformation within the transformation layer, thereby circumventing the need for computing the inner products. Consequently, the computation complexity of our attention mechanism in the transformation layer is reduced to $\mathcal{O}(N_m^2)$, representing a reduction factor of $D$. Additionally, since our method calculates the query embeddings solely within the \less, our K-Q-V transformation complexity is likewise diminished by a factor of $3$. 

In the downsampling layer between consecutive stages, considering a downsample rate of $2$ as an example, the computational complexity of the DWConv in the attention downsampling layer can be calculated as $\textrm{Complexity} = 2 \times 2 \times \frac{N_m}{2} \times \frac{N_m}{2} \times D = \mathcal{O}(N_m^2D)$. Similarly, the complexity of the $\textrm{Conv}_{1\times1}$ operation in the attention residual module is also $\mathcal{O}(N_m^2D)$. However, it is important to note that attention downsampling only occurs once per stage. Therefore, the additional complexity introduced by these operations is negligible when compared to the complexity reduction achieved by the \less.

%% file: sec/4Exp.tex
\section{Experiments}
\label{sec:experiments}
In this section, we evaluate our model's performance on two benchmark datasets: ImageNet-1K \cite{krizhevsky2017imagenet} for classification, COCO2017 \cite{COCO} for detection and ADE20K \cite{ADE20K} for segmentation. We compare our model with other state-of-the-art works on these datasets to demonstrate its effectiveness and efficiency. Furthermore, we perform ablation studies to investigate the necessity and contributions of each component in the proposed model. This analysis provides valuable insights into the role of each part and helps to establish the efficacy of our approach.


\begin{table}[t]
\centering
\caption{Detailed configurations of the LaViT series. 'Blocks' and 'Heads' refer to the number of blocks ($[L^1, L^2, L^3, L^4]$) and heads in four stages, respectively.  'Channels' refers to the input channel dimensions across the four stages. And '$N_\text{LA}$' denotes the layer within each stage at which the utilization of the \less begins.}
\label{table:configurations}
\resizebox{0.47\textwidth}{!}{
\begin{tabular}{@{}c|c|c|c|c@{}}
\toprule
Models        & Channels & Blocks       & Heads        & $N_\textrm{LA}$                                                    \\ \midrule
LaViT-T & {[}64,128,320,512{]}         & {[}2,2,2,2{]} & {[}1,2,5,8{]}  & {[}0,0,2,2{]}  \\
LaViT-S & {[}64,128,320,512{]}         & {[}3,4,6,3{]} & {[}1,2,5,8{]} & {[}0,0,3,2{]}   \\
LaViT-B & {[}64,128,320,512{]}        & {[}3,3,18,3{]} & {[}1,2,5,8{]} & {[}0,2,4,3{]}   \\ \bottomrule
\end{tabular}
}
\end{table}

\begin{table}[t]
\vspace{-20pt}
\centering
\resizebox{0.95\linewidth}{!}{
\begin{tabular}{l|c|c|c|c}
\toprule
\multirow{2}{*}{Model} & ~Params~ & ~~FLOPs~~ & ~~Throughput~~ & ~~Top1~~ \\
                       & (M)     & (G)    & (image/s) & (\%) \\
\midrule
ResNet-18    &   11.7         &     ~~1.8      &  \textbf{\color{blue} 4454} &   69.8      \\
RegNetY-1.6G &     11.2       &   ~~1.6        &   1845 &   78.0   \\
DeiT-T      &     \textbf{\color{blue}~~5.7}       &     \textbf{\color{blue}~~1.3}   & 3398   &     72.2    \\
PVT-T       &       13.2     &   ~~1.9  & 1768       &   75.1      \\
PVTv2-b1       &       13.1     &   ~~2.1      & 1231  &   78.7      \\
\midrule
LaViT-T  &       10.9     &   ~~1.6        & 2098 &    \textbf{79.2}   \\
\midrule
ResNet-50    &       25.0     &     ~~4.1      &  1279 & 76.2      \\
RegNetY-4G &       20.6     &  ~~4.0         &  1045 &  79.4     \\
EfficientNet-B4 &       \textbf{\color{blue}19.0}    &  ~~4.2        & 387 &   82.4     \\
EfficientViT-B2 &       24.0    &  ~~4.5        & 1587 &   82.1     \\
DeiT-S         &      22.1      & ~~4.6      & 1551    & 79.9        \\
DeepViT-S & 27.0 &~~6.2 & 1423 &82.3 \\
PVT-S      &      24.5      &     ~~3.8      &  1007 &  79.8      \\
CvT-S & 25.8 & ~~7.1 & 636& 82.0 \\
Swin-T&      28.3      &   ~~4.5        & 961 &   81.2     \\
PVTv2-b2     &    25.4      &      ~~4.0    & 695   &    {82.0}      \\
DynamicViT-S (90\%) &    24.1      &      ~~4.0    & 1524   &    {79.8} \\
EViT-S (90\%) &    23.9      &      ~~4.1      & \textbf{\color{blue} 1706} &    {79.7} \\
LiT-S  &   27.0      &      ~~4.1     & 1298  &    {81.5} \\
PPT-S  &    22.1      &      ~~3.1    & 1698   &    {79.8} \\

\midrule
LaViT-S &        22.4    &      \textbf{\color{blue}~~3.3}     & 1546 &    \textbf{82.6}    \\
\midrule
ResNet-101      &     45.0     &    ~~7.9     & 722 & 77.4     \\
ViT-B &     86.6     &    17.6     &   270&77.9     \\
DeiT-B  &     86.6     &     17.5    &582 &  81.8    \\
Swin-S &      49.6     &   ~~8.7   & 582&   83.1     \\
Swin-B &      87.8    &   15.4  &386 &   \textbf{83.4}     \\
DynamicViT-B (90\%) &    76.6      &      ~~14.1    & 732   &    {81.5} \\
EViT-B (90\%) &    78.6      &      ~~15.3      & 852 &    {81.3} \\
LiT-M  &   48.0      &      ~~8.6     & 638  &    {83.0} \\
PPT-B &    86.0      &      ~~14.5    & 714   &    {81.4} \\
PVT-M  &     44.2     &     ~~6.7  & 680  &   81.2   \\
PVT-L &     61.4    &    ~~9.8  & 481&  81.7   \\

\midrule

LaViT-B       &     \textbf{\color{blue}39.6}   &    \textbf{\color{blue}~~6.1}    & \textbf{\color{blue} 877}  &   {83.1}      \\
\bottomrule
\end{tabular}}
\vspace{2pt}
\caption{Comparison of different backbones on ImageNet-1K classification. Except for EfficientNet (EfficientNet-B4), all models are trained and evaluated with an input size of $224\times 224$. The least computations and fastest throughput appear in \textbf{\color{blue} blue bold}, and the best results appear in \textbf{bold}.\protect\footnotemark[1]}
\label{classification}
\end{table}
\footnotetext[1]{The 90\% notation in brackets indicates that we keep the token ratio of 90\% to represent the visual data during the training of the corresponding ViTs---DynamicViT and EviT, respectively. Additionally, given our aim to strike a balance between efficiency and effectiveness, we will not compare our results to high-performance but computationally intensive models, such as Swin-B-V2~\cite{liu2022swin} and ConvNeXt-B~\cite{liu2022convnet}.}

\subsection{Architecture Variants}
To ensure an equitable comparison with other models while maintaining a similar level of computational complexity, we establish three models: \name-T, \name-S, and \name-B.  The detailed configuration information is provided in Table~\ref{table:configurations}, and we follow the same network structure as PVT~\cite{wang2021pyramid, wang2022pvt} except for introducing the {Less-Attention} Transformer encoder and skip-connection attention downsampling.
The number of blocks, channels, and heads affects the computational cost.

\subsection{Baselines}
We conduct a thorough experimental evaluation of our proposed method by comparing it with various CNNs, ViTs, and hierarchical ViTs.  Specifically, the following baselines are used:
\begin{itemize}
    \item {\bf CNNs:} ResNet~\cite{he2016deep},  RegNet~\cite{regnet} and EfficientNet~\cite{tan2019efficientnet}.
    \item {\bf ViTs:} ViT~\cite{ViT}, DeiT~\cite{touvron2021training}, CvT~\cite{yuan2021incorporating},DeepViT~\cite{zhou2021deepvit},  FocalViT~\cite{yang2021focal} and SwinTransformer~\cite{Swin}.
    \item  {\bf Efficient ViTs:} HVT~\cite{pan2021hvt}, PVT~\cite{wang2021pyramid}, DynamicViT~\cite{dynamicvit}, EViT~\cite{Evit}, LiT~\cite{pan2022less}, EfficientViT~\cite{cai2022efficientvit} and PPT~\cite{wu2023ppt}.
\end{itemize}

\begin{table*}[htbp]
\vspace{-40pt}
\centering
\resizebox{0.80\textwidth}{!}{
\begin{tabular}{@{}l|cc|cccccc|cccccc@{}}
\toprule
\multirow{2}{*}{Backbone}    & \#Param.      & FLOPs      & \multicolumn{6}{c|}{RetinaNet 1$\times$}      & \multicolumn{6}{c}{RetinaNet 3$\times$ + MS}  \\ 
\cmidrule(l){2-15}  & (M)        & (G)        & AP$^b$            & AP$_{50}^b$           & AP$_{75}^b$           & AP$_S^b$           & AP$_M^b$            & \multicolumn{1}{c|}{AP$_L^b$}            & AP$^b$            & AP$_{50}^b$            & AP$_{75}^b$            & AP$_S^b$            & AP$_M^b$           & AP$_L^b$            \\ \midrule
ResNet50                   & 38       & 239        & 36.3          & 55.3          & 38.6          & 19.3          & 40.0          & \multicolumn{1}{c|}{48.8}          & 39.0          & 58.4          & 41.8          & 22.4          & 42.8          & 51.6          \\
PVT-Small                  & 34       & 226        & 40.4          & 61.3          & 43.0          & 25.0          & 42.9          & \multicolumn{1}{c|}{55.7}          & 42.2          & 62.7          & 45.0          & 26.2          & 45.2          & 57.2          \\
Swin-T                       & 39       & 245        & 41.5          & 62.1          & 44.2          & 25.1          & 44.9          & \multicolumn{1}{c|}{55.5}          & 43.9          & 64.8          & 47.1          & 28.4          & 47.2          & 57.8          \\
\textbf{LaViT-T(ours)} & 33 & 202 & \textbf{46.2} & \textbf{67.2} & \textbf{49.1} & \textbf{29.6} & \textbf{50.2} & \multicolumn{1}{c|}{\textbf{61.3}} & \textbf{48.4} & \textbf{69.9} & \textbf{51.7} & \textbf{31.8} & \textbf{52.2} & \textbf{64.1} \\ \midrule
ResNet101                    & 58       & 315        & 38.5          & 57.8          & 41.2          & 21.4          & 42.6          & \multicolumn{1}{c|}{51.1}          & 40.9          & 60.1          & 44.0          & 23.7          & 45.0          & 53.8          \\
PVT-M                 & 54       & 283        & 41.9          & 63.1          & 44.3          & 25.0          & 44.9          & \multicolumn{1}{c|}{57.6}          & 43.2          & 63.8          & 46.1          & 27.3          & 46.3          & 58.9          \\
Swin-S            & 60       & 335        & 44.5          & 65.7          & 47.5          & 27.4          & 48.0          & \multicolumn{1}{c|}{59.9}          & 46.3          & 67.4          & 49.8          & 31.1          & 50.3          & 60.9          \\
\textbf{LaViT-S(ours)} & 47 & 290 & \textbf{46.7} & \textbf{68.3} & \textbf{49.7} & \textbf{29.9} & \textbf{50.7} & \multicolumn{1}{c|}{\textbf{61.7}} & \textbf{48.9} & \textbf{70.3} & \textbf{52.2} & \textbf{33.1} & \textbf{52.6} & \textbf{65.4}  \\ 
\bottomrule
\end{tabular}}
\caption{Results on COCO object detection using the RetinaNet~\cite{lin2017focal} framework. 1$\times$ refers to 12 epochs, and 3$\times$ refers to 36 epochs. MS means multi-scale training. AP$^b$ and AP$^m$ denotes box mAP and mask mAP, respectively. FLOPs are measured at resolution $800 \times 1280$.}
\vspace{-10pt}
\label{table:result_COCO}
\end{table*}
\subsection{Image Classification on ImageNet-1K}
\label{sec:classification}
\noindent \textbf{Settings.}
The image classification experiments are conducted on the ImageNet-1K dataset. Our experimental protocol follows the procedures outlined in DeiT \cite{touvron2021training}, with the exception of the model itself. Specifically, we apply the same data augmentation and regularization techniques employed in DeiT.
We utilize the AdamW optimizer \cite{AdamW} to train our models from scratch for 300 epochs (with a 5-epoch warm-up). The initial learning rate is set to 0.005 and varies according to a cosine scheduler. The global batch size is set to 1024, distributed across 4 GTX-3090 GPUs. During the test on the validation set, the input images are first resized to 256 pixels, followed by a center crop of 224 x 224 pixels to evaluate the classification accuracy.

\noindent \textbf{Results.}
We present the classification results on ImageNet-1K in Table \ref{classification}. The models are classified into three groups based on their computational complexity: tiny (approximately 2G), small (approximately 4G), and base (approximately 9G). Our approach achieves competitive performance compared to state-of-the-art ViTs with markedly reduced computational requirements. Specifically, in the tiny and small model clusters, our method surpasses all other existing models by at least 0.2\% and 0.5\%, respectively, while maintaining a substantially lower computational cost, which is our principal concern. In the base-size models, our architecture, which incorporates the base structure of PVT but includes the Less-Attention component, demonstrates superior performance over two PVT-based models (PVT-M and PVT-L).  Furthermore, we also compare our architecture to several efficient ViT designs (DynamicViT, EViT, LiT, efficientViT and PPT). We observe that our results reflect a better balance between effectiveness and efficiency. Note that our design necessitates reduced computation cost owing to our resource-efficient Less-Attention mechanism, rendering our lightweight module an attractive option for implementing ViT on mobile platforms.

\subsection{Object Detection on COCO2017}
\noindent \textbf{Settings.}
We conduct the detection experiments on COCO 2017~\cite{COCO} dataset. We test the model effectiveness on RetinaNet~\cite{lin2017focal}. We follow the common practice by initializing the backbone with pre-trained weights obtained from ImageNet-1K. In addition, we use AdamW~\cite{AdamW} optimizer, and train the network with the batchsize of 16 on 8 GPUs.

\noindent \textbf{Results.} \quad
We present the results of object detection in Table~\ref{table:result_COCO}. It is evident that our LaViT model exhibits a notable advantage over both its CNN and Transformer counterparts. Specifically, with the $1\times$ schedule, our tiny version LaViT-T achieves 9.9-12.5 $\text{AP}^b$ against ResNet under comparable settings, while the small version LaViT-S outperforms its CNN counterpart by 8.1-10.3 $\text{AP}^b$. This trend persists with the $3\times$ schedule, as our LaViT consistently demonstrates competitive performance. Particularly noteworthy is our architecture's ability to consistently outperform the Swin Transformer in terms of detection performance while imposing a smaller training burden. Thus, the results on COCO2017 reaffirm our assertion that our carefully designed LaViT model enhances feature extraction with reduced computational overhead.

\begin{table*}[t]
\centering
\vspace{-40pt}
\resizebox{0.75\linewidth}{!}
{
\small
\begin{tabular}{l|ccc|cccc}
\toprule
\multirow{2}{*}{Backbone} & \multicolumn{3}{c|}{Semantic FPN 80k} & \multicolumn{4}{c}{UperNet 160K}       \\
                          & Param (M)    & FLOPs (G)    & mIOU (\%)   & Param (M) & FLOPs (G) & mIOU (\%) & MS mIOU (\%)\\
\midrule
ResNet-50 &       28.5       &    183          &  36.7 &  -     &       -    &   -   &   - \\
Swin-T   &    31.9     &     182    &    41.5  &    59.9    &      945     & 44.5     &  45.8  \\
PVT-S &30.2      &    146    & 43.2   &     -    &   - & -  & -  \\
Twin-S      &   28.3   &  144         &43.2&54.4           & 932          &46.2      &47.1   \\
LiT-S      &   32.0   &  172         &41.3&57.8           & 978          &44.6      &45.9   \\
Focal-T    &    -          &      -   &   -     &62.0           &998           &45.8      &   47.0 \\
LaViT-S    &       \textbf{\color{blue}25.1}       &     \textbf{\color{blue} 122}    &  \textbf{44.1}     & \textbf{\color{blue}52.0}            &\textbf{\color{blue}920}           &\textbf{47.2}      &   \textbf{49.5}  \\
\bottomrule 
\end{tabular}}
\vspace{2pt}
\caption{Segmentation performance of different backbones in Semantic FPN and UpperNet framework on ADE20K. The least computation appears in \textbf{\color{blue} blue bold}, and the best results appear in \textbf{ bold}.}
\label{seg}
\end{table*}

\begin{table}[t!]
    \centering
    \resizebox{1\linewidth}{!}{
    \begin{tabular}{l|ll|ll}
    \toprule
    \multirow{2}{*}{Backbone} & \multicolumn{2}{c|}{Tiny} & \multicolumn{2}{c}{Small} \\
    & Top-1 Acc(\%)    & FLOPs (G)    & Top-1 Acc(\%)    & FLOPs (G) \\
    \midrule
    ViT & 72.2 & 1.4 & 79.1 & 4.6 \\
    ViT$_\textrm{+LA}$ & $73.2({\color{blue} \uparrow 1.0})$ & $1.2({\color{red} \downarrow 14.2\%})$ & $80.0({\color{blue} \uparrow 0.9})$ & $4.0({\color{red} \downarrow 13.1\%})$ \\
    \midrule
    DeiT & 72.2 & 1.4 & 79.9 & 4.7 \\
    DeiT$_\textrm{+LA}$ & $73.4({\color{blue} \uparrow 1.2})$ & $1.2({\color{red} \downarrow 14.2\%})$ & $80.4({\color{blue} \uparrow 0.5})$ & $4.2({\color{red} \downarrow 10.6\%})$ \\
    \midrule
    DeepViT & 73.4 & 1.5 & 80.9 & 4.8 \\
    DeepViT$_\textrm{+LA}$ & $73.8({\color{blue} \uparrow 0.4})$ & $1.1({\color{red} \downarrow 25.8\%})$ & $81.4({\color{blue} \uparrow 0.5})$ & $4.2({\color{red} \downarrow 12.6\%})$ \\
    \midrule
    CeiT & 76.2 & 1.2 & 82.0 & 4.5 \\
    CeiT$_\textrm{+LA}$ & $76.7({\color{blue} \uparrow 0.5})$ & $1.1({\color{red} \downarrow 9.0\%})$ & $82.4({\color{blue} \uparrow 0.4})$ & $4.1({\color{red} \downarrow 8.8\%})$ \\
    \midrule
    HVT & 75.7 & 1.4 & 80.4 & 4.6 \\
    HVT$_\textrm{+LA}$ & $76.2({\color{blue} \uparrow 0.5})$ & $1.2({\color{red} \downarrow 15.2\%})$ & $80.8({\color{blue} \uparrow 0.4})$ & $4.2({\color{red} \downarrow 13.4\%})$ \\
    \midrule
    PVT & 75.1 & 1.9 & 79.8 & 3.8 \\
    PVT$_\textrm{+LA}$ & $75.9({\color{blue} \uparrow 0.8})$ & $1.4({\color{red} \downarrow 25.6\%})$ & $80.4({\color{blue} \uparrow 0.6})$ & $3.2({\color{red} \downarrow 15.7\%})$ \\
    \midrule
    Swin & 81.2 & 4.5 & 83.2 & 8.7 \\
    Swin$_\textrm{+LA}$ & $81.7({\color{blue} \uparrow 0.5})$ & $4.0({\color{red} \downarrow 11.1\%})$ & $83.5({\color{blue} \uparrow 0.3})$ & $7.8({\color{red} \downarrow 10.3\%})$ \\
    
    \bottomrule
    \end{tabular}}
    \vspace{2pt}
    \caption{Top-1 classification accuracy on ImageNet-1k using different transformer backbones and their corresponding Less-Attention plug-in variants. Footnote 'LA' indicates the addition of our Less-Attention module to the respective backbone architectures. $\color{blue} \uparrow$ and $\color{red} \downarrow$ denote the increase in Top-1 accuracy and the percentage of FLOPs reduction, respectively.}
    \label{tab:extendability}
\end{table}

\subsection{Semantic Segmentation on ADE20K}
\noindent \textbf{Settings.}\quad
We conduct experiments on semantic segmentation using the ADE20K dataset, which comprises 150 classes and 20,000 images for training, and 2,000 images for validation. Our backbone networks for segmentation are Semantic FPN \cite{lin2017feature} and UperNet \cite{xiao2018unified}. We follow the training settings established in \cite{Swin} and resize images to $512\times512$ for training. 
We train  UperNet for 160k iterations and SemanticFPN for 80k iterations. The initial learning rate is set to $6\times10^{-5}$, utilizing a poly scheduler for learning rate decay. The experiment is conducted by using the batch size of 16 across $4$ GTX3090 GPUs.

\noindent \textbf{Results.} \quad
Table \ref{seg} provides an overview of the segmentation results. Our model demonstrates superiority over Swin Transformer, exhibiting an mIoU improvement of +2.6 with Semantic FPN and +2.7 with UperNet. In the Semantic FPN test, our LaViT-S achieves a relatively modest increase of +0.9 mIoU compared to the baseline (PVT-S), but notably with significantly fewer computations. When integrated into the UperNet architecture, LaViT achieves substantial improvements of +2.7 mIoU, +1.0 mIoU, and +1.4 mIoU when compared to various mainstream models. These competitive results are maintained even when employing test time augmentation. In particular, LaViT-S outperforms Focal-T by +1.4 mIoU and +2.5 MS mIOU. These findings underscore LaViT's ability to produce high-quality semantic segmentation outputs while operating within the framework of its computation-efficient attention mechanism.

\begin{figure} [t]
	\centering
	\subfloat[ViT]{
		\includegraphics[scale=0.48]{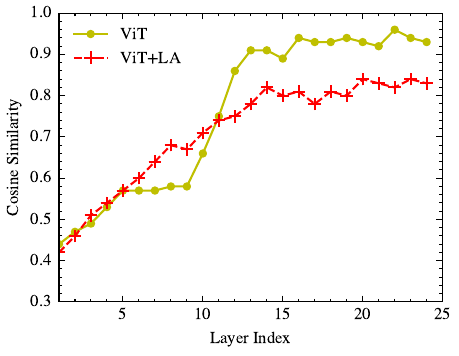}\label{vit}}
	\subfloat[PVT]{
		\includegraphics[scale=0.48]{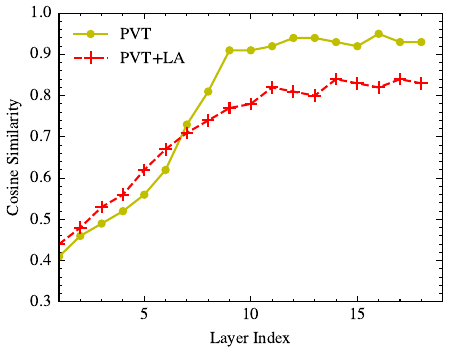}\label{pvt}
		}
	\caption{The similarity ratio of the generated self-attention maps of the current layer with its previous layer. }
	\label{fig: visualization} 
\end{figure}

\subsection{Ablation Study}

\noindent \textbf{Attention Saturation.} 
To demonstrate the efficacy of our Less-Attention module in addressing attention saturation, we present the attention similarity ratio (cosine similarity computed by the attention map in the current layer and its previous layer) in \Cref{fig: visualization}. We conduct the comparison using two backbones, namely, ViT and PVT. In \ref{vit}, we select the ViT architecture with 25 layers and no hierarchical structure. In \ref{pvt}, we employ PVT-M as our baseline and assess the attention similarity at the 3rd stage, which consists of 18 layers. Both sets of results clearly illustrate that the original architecture encounters a significant attention saturation issue. However, this phenomenon is effectively mitigated by incorporating our modules, enabling deep attention to fulfill its intended role. \\

\noindent \textbf{Extendability of Less-Attention Module.} We extend our Less-Attention module to various ViT architectures, and report the results in Table \ref{tab:extendability}. 
The incorporation of the \less into any of the foundational Transformer architectures leads to enhancements in accuracy while concurrently reducing computational demands. Notably, the most significant improvement is observed when incorporating the module into the vanilla ViT/DeiT architecture. This may be attributed to the fact that the vanilla ViT/DeiT 
does not have a hierarchical structure, thereby experiencing considerable attention saturation issues. Moreover, when integrating our method into DeepViT, we observe the most substantial decrease in computational resources. These findings jointly underscore the scalability of our method, demonstrating that the application of LA module can render existing ViT architectures more practical and feasible. \\

\begin{table}[t]
	\centering
 \vspace{-10pt}
	\resizebox{0.8\linewidth}{!}{
		\begin{tabular}{lccccll}  
			\toprule
			\multirow{2}{*}{Model}& \multicolumn{3}{c}{Module}& \multirow{2}{*}{Tiny} & \multirow{2}{*}{Small}\\  
			& AR & LA & $\mathcal{L}_\textrm{DP}$  & & \\
			\midrule
			w/o LA &-&-&-&78.7& 82.0 \\
			w/o AR &-&\checkmark&\checkmark&79.0&82.2 \\
			LaViT&\checkmark&\checkmark&\checkmark&79.2&82.6\\
			w/o $\mathcal{L}_\textrm{DP}$&\checkmark&\checkmark&- &$59.1({\color{red} \downarrow 20.1})$&$57.1({\color{red} \downarrow 25.5})$\\
			\bottomrule
	\end{tabular}}
 \vspace{2pt}
	\caption{Ablation study of the proposed module on the ImageNet-1k dataset. Baseline means we {remove} all the proposed modules, resulting in the PVT Transformer baseline. 'AR' and 'LA' indicate the Attention-Residual and Less-Attention modules, respectively. 'w/o $\mathcal{L}_\text{DP}$' indicates we remove the Diagonality Preserving loss.}
	\label{tbl:ablation}
\end{table}

\noindent \textbf{Importance of Each Component.} 
We conduct ablation studies on the proposed module with the ImageNet-1k dataset, and the results are shown in Table \ref{tbl:ablation}. On both networks (\ie, tiny and small), our proposed modules prove to be indispensible for Transformer training. The baseline, which replaces the \less with MHSA, corresponds exactly to the PVT model, exhibiting a decrease in predictive accuracy by 0.5\% and 0.6\% compared to our model. Additionally, removing the attention residual modules, denoted as ``w/o AR'', results in a reduction of predictive accuracy by 0.2\% and 0.4\%. Lastly, and most importantly, we assert that the additional loss function to preserve diagonal correlations is vital for effectively comprehending semantic information in visual data. When relying solely on the CE loss, the model's predictions deteriorate. This might be attributed to the potential limitation of relying solely on the transformation for attention matrices, which could compromise their capacity to express correlations among tokens. All these experimental findings collectively emphasize the contribution of each component within our model architecture. \\

\begin{table}[]
\vspace{-20pt}
    \centering
    \resizebox{0.75\linewidth}{!}{
    \begin{tabular}{c|ccc|ccc}
    \toprule
    \multirow{2}{*}{Model} & \multicolumn{3}{c|}{Stage 3}& \multicolumn{3}{c}{Stage 4} \\
    & L2 & L3 & L4 & L1 & L2 & L3 \\
    \midrule
    LaViT-S & 79.1 & 82.6 & 82.4 & 80.1 & 82.6 & 82.3 \\
    \midrule
    LaViT-B & 78.9 & 82.5 & 83.1 & 80.4 & 82.3 & 83.1 \\
    \bottomrule
    \end{tabular}}
    \caption{Ablation study on the layer where Less-Attention starts. We conduct experiments on the last two stages---Stage 3,4. 'L2' beneath Stage 3 means we use the \less to replace the vanilla encoder from the second layer in the third Stage.}
    \label{tab:start layer}
    \vspace{-15pt}
\end{table}

\noindent \textbf{Less-Attention Selection.} In deep ViTs, careful selection of the starting layer for Less-Attention is crucial. Thus, we design experiments to  select the starting layer for Less-Attention in the network architecture, and the results are presented in Table~\ref{tab:start layer}. 
As shown in the table, 
directly using the Less-Attention layer from the second layer in the stage leads to a decrease in model performance. This phenomenon could be attributed to overly relying on the semantics of the first MHSA layer. Thus, leveraging the \less at deeper layers in the stage may  mitigate this issue. Furthermore, while utilizing the \less at relatively deeper layers does not affect the model performance much, it may lead to increased  computational costs. This contradicts the design objective of our architecture to reduce the computational overhead.

%% file: sec/5Con.tex
\section{Conclusion}
Aiming to reduce the costly self-attention computations, we proposed a new model called {Less-Attention} Vision Transformer (\name). \name leverages the computed dependency in Multi-Head Self-Attention (MHSA) blocks and bypasses the attention computation by
re-using attentions from previous MSA blocks. We additionally incorporated a straightforward Diagonality Preserving loss, designed to promote the intended behavior of the attention matrix in representing relationships among tokens. Notably, our Transformer architecture effectively captures cross-token associations, surpassing the performance of the baseline while maintaining a computationally efficient profile in terms of 
quantity of parameters and floating-point operations per second (FLOPs). 
Comprehensive experimentation has confirmed the efficacy of our model as a foundational architecture for multiple downstream tasks. Specifically, the proposed model demonstrates superiority over previous Transformer architectures, resulting in state-of-the-art performance in classification and segmentation tasks. 